\documentclass[11pt]{article}
\usepackage[utf8]{inputenc}
\usepackage{geometry}
\usepackage{graphicx}
\usepackage{amsmath, amssymb}
\usepackage{hyperref}
\usepackage{authblk}
\usepackage{setspace}
\usepackage{csquotes}
\usepackage[super,sort&compress]{natbib}
\usepackage{booktabs}

\title{War in the Abstract: The Rise and Consequences of Militarized Language in Scientific Communication}

\author[1]{Sovesh Mohapatra}
\author[2]{David Lydon-Staley}
\author[1,3,4,5]{Dani S. Bassett}
\affil[1]{Department of Bioengineering, School of Engineering and Applied Science, University of Pennsylvania, PA, USA}
\affil[2]{Annenberg School of Communication, University of Pennsylvania, PA, USA}
\affil[3]{Departments of Electrical \& Systems Engineering and Physics \& Astronomy, University of Pennsylvania, PA, USA}
\affil[4]{Departments of Psychology and Biomedical Engineering, Yale University, CT, USA}
\affil[5]{Wu Tsai Institute, Yale University, CT, USA}
\date{}

\begin{document}

\maketitle

\pagebreak
\section*{Abstract}
\noindent
Scientists do not, by profession, wage war. Yet warfare's vocabulary consistently appears in their abstracts. To quantify the extent to which warfare's vocabulary pervades scientific abstracts, we analyze 21.4 million papers (2010-2025; OpenAlex, PubMed). We additionally run a within-subject war-framing experiment ($N = 801$; 32{,}040 trials) designed to provide causal insight into the effects of militaristic language on persuasion. Between 2010 and 2025, the presence of militaristic terms in scientific abstracts rose 48\% in OpenAlex and 32\% in PubMed, with the rise accelerating sharply after 2019 (cross-database $r = 0.96$, $p < 10^{-8}$). The prevalence of militaristic language is conflict-aligned at both country and annual scales (Uppsala Conflict Data Program; $r = 0.77$-$0.84$), with the abstracts from the Global South displaying the fastest rise in militaristic language. Among disciplines, social sciences leads in level of such language while engineering and computer science lead in growth. The COVID and post-2022 large-language-model eras also accelerated the rise and narrowed the language gap between native-English-speaking and non-English-speaking countries. In our follow-up experiment, we found that war framing reduced credibility (mean shift $-0.176$ Likert units, 95\% CI $[-0.21, -0.14]$; $d_z = -0.28$, $p < 10^{-20}$), funding willingness ($d_z = -0.12$) and policy support ($d_z = -0.08$), with a trend-level increase in sense of urgency ($d_z = +0.07$). Collectively, findings reveal that while scientific abstracts drift toward warfare, the use of militaristic language may erode credibility, funding willingness, and policy support.

\pagebreak

\noindent
Scientists do not wage war, yet they have come to write as if they did. They speak of ``wars'' on cancer and dementia, of immune cells that ``patrol'' the body for ``invaders,'' of ``magic bullets'' against tumors, and of ``arms races'' between hosts and pathogens. Intuitively, war metaphors may persist in scientific writing because authors perceive them as effective rhetorical instruments\cite{sopory2002persuasive}. They can make abstract targets concrete\cite{lakoff1980metaphors,landau2010metaphor}, signal urgency and stakes\cite{flusberg2017metaphors}, and have been deployed historically to mobilize public attention and policy support around scientific priorities\cite{penson2004cancer,thibodeau2011metaphors}. Despite perceptions that militaristic language will be persuasive, there is increasing evidence that the use of war metaphors may undermine message effectiveness. In her 1978 critique, Sontag \cite{sontag1978illness} argued that the framing of cancer as an enemy to be `fought' and patients as `fighters' imposes a moral burden on the ill, stigmatizing those who fail to recover. Subsequent experimental work has supported Sontag's concerns: war framings of cancer reduce engagement with prevention\cite{hauser2015war} and increase patient guilt\cite{hendricks2018emotional}. Parallel effects of war framings have been observed across other domains: in crime, they shift policy preferences toward punitive enforcement\cite{thibodeau2011metaphors}; in climate change, they narrow public discourse to a binary win-or-lose frame\cite{flusberg2017metaphors}; and in pandemic response, they encourage blame and stigmatization of outgroups\cite{wicke2020framing,semino2021fire}. Despite the unintended effects of war framings in scientific work, prior scholarship has documented persistent war metaphors in immunology\cite{martin1994flexible}, oncology\cite{penson2004cancer} and microbiology\cite{board2006ending}. None of these prior accounts measures the trajectory of war language across the broader scientific corpus, the recent escalations of armed conflict, or the post-2022 emergence of large language models. Given increasing reason to believe that the use of war-related language will reduce credibility, it is an important moment to examine whether scientific writing has tilted further toward this vocabulary or held steady against decades of critique remains unmeasured\cite{intemann2023sciencecomm,morenocastro2026reshaping}.

The prevalence of war metaphors in scientific work is consequential because metaphor acts on the reader, not just on the text\cite{landau2010metaphor}. Metaphors take a familiar source domain (such as warfare) and project its structure onto an abstract target (such as a disease) \cite{lakoff1980metaphors,gibbs1994poetics}, and the choice of source domain produces measurable shifts in reasoning about the target\cite{boroditsky2001does}. These shifts operate largely outside the reader's awareness\cite{bargh1999automaticity}: experimental participants attribute their conclusions to underlying evidence rather than to the metaphor\cite{thibodeau2011metaphors}. These framing effects persist even when the frame is explicitly identified, complicating attempts to debias readers through warnings alone\cite{chong2007framing}. Across the experimental literature, metaphor produces consistent persuasive effects on attitudes and beliefs, with meta-analytic evidence supporting the generality of the mechanism\cite{sopory2002persuasive}. Metaphor's effects do not require deliberate rhetorical use; conventional metaphors woven into routine vocabulary shape interpretation as effectively as deliberate ones\cite{steen2008paradox}. War metaphors, in particular, carry a cognitive profile distinct from other source domains. They prime competition, urgency, and a binary frame of outcomes as wins or losses\cite{flusberg2017metaphors}. They elevate the perceived competence of a speaker while depressing the perceived warmth\cite{fiske2014gaining}. In lay-audience research, perceived warmth is among the strongest predictors of whether information from an expert source is accepted as legitimate\cite{hendriks2015meti}. These individual-experiment effects are modest; yet, compounded across millions of scientific texts, the cumulative consequences for lay reasoning are substantial.

Three convergent shifts in the 2010s and 2020s intersect with the trajectory of war framing in science, making the timing of any rise analytically central. First, armed conflict has escalated globally since 2014\cite{pettersson2020organized}, raising the importance of literal warfare in public discourse. Second, the COVID-19 pandemic was widely framed as a war in both public-health and social-media discourse\cite{wicke2020framing,semino2021fire}, and has been followed by a measurable decline in public trust in scientific institutions that raises the stakes of framing\cite{intemann2023sciencecomm,morenocastro2026reshaping}. Third, the post-2022 emergence of large language models\cite{bommasani2021opportunities} has begun to alter the lexical register of scientific writing\cite{liang2025quantifying,kobak2025delving,geng2024chatgpt}. These drivers operate at different levels: conflict and pandemic framing shift the discursive environment in which scientists work, while large language models (LLMs) shift the surface form of the writing itself. Whether scientists deliberately import the war frame or absorb it from the surrounding discourse raises an attribution question that prevalence measurement cannot fully resolve. Each driver nonetheless has a distinct temporal signature that timing analysis can exploit.

No prior work has measured the trajectory of war vocabulary in science at corpus scale. The bibliometric tradition of computational science-of-science\cite{foster2015tradition,liu2023data} tracks topical drift, citation patterns and authorship dynamics, but rarely the rhetorical register of scientific writing. Recent population-scale lexical work has focused on LLM-marker vocabulary\cite{liang2025quantifying,kobak2025delving}, sentiment in news consumption\cite{robertson2023negativity} and shifts in positivity, hype and readability across scientific abstracts\cite{vinkers2015use,millar2019hype,plaven2017readability}, not on the prevalence of war terms in scientific writing itself. Such measurements face a further methodological caution: lexicon-prevalence findings can be artifacts of unrelated text statistics or of broader semantic drift rather than substantive signals\cite{burton2021reconsidering,hamilton2016diachronic,kulkarni2015statistically}. Any rising-lexicon claim must therefore clear three methodological bars, which our design meets: cross-corpus replication, placebo benchmarking, and robustness across publication-composition and language shifts.

\begin{figure}[!ht]
  \centering
  \includegraphics[width=\textwidth]{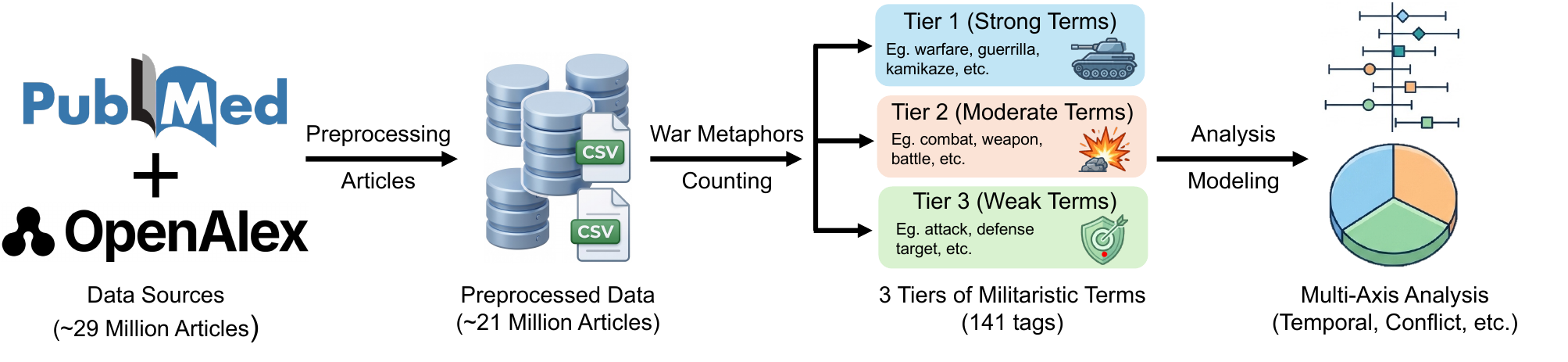}
  \caption{\textbf{Study design and analytical pipeline for quantifying militarized language in scientific publications.}
  We assembled a corpus of approximately 29~million peer-reviewed articles published between 2010 and 2025 from two complementary bibliographic databases: OpenAlex (multidisciplinary) and PubMed (biomedical). After harmonization, deduplication and removal of records lacking abstracts or titles, $N \approx 21.4$~million papers were retained for downstream analysis. Title and abstract text were tokenized and screened against a hand-curated lexicon of 141 militaristic terms, organized into three precision-stratified tiers. \textbf{Tier~1} (T1; e.g.\ warfare, guerrilla, kamikaze) captures direct, unambiguous military vocabulary. \textbf{Tier~2} (T2; e.g.\ combat, weapon, battle) captures moderately martial terms whose figurative use within scientific discourse is interpretable but not fully ambiguous. \textbf{Tier~3} (T3; e.g.\ attack, defense, target) captures high-frequency, dual-use vocabulary whose military reading is context-dependent (the complete term list for all three tiers is given in Supplementary Table~S2). The per-paper tier indicators were aggregated into yearly prevalence series and projected along five orthogonal analytical axes: temporal dynamics and structural breakpoints; geographic and hemispheric variation; disciplinary stratification; association with armed conflict using UCDP/PRIO data; and other targeted analyses including pandemic, language, and large-language-model-era effects.}
  \label{fig:pipeline}
\end{figure}

Here, we analyze 21.4 million papers from two disjoint bibliographic databases (OpenAlex and PubMed) published between 2010 and 2025 (Fig.~\ref{fig:pipeline}). A three-tier lexicon separates literal military terms from metaphorical and ambiguous usages, providing a diagnostic for the underlying mechanism: a rise driven by literal warfare would concentrate in Tier 1, whereas a rise driven by metaphorical adoption would concentrate in Tiers 2 and 3. A frequency-matched placebo lexicon controls against generic text-statistics artifacts. The cross-database trajectory is assessed for structural breakpoints using changepoint detection. The country-level prevalence is mapped across 91 nations and tested against the Uppsala Conflict Data Program counts at both annual and country scales. Disciplinary stratification across eight super-disciplines decomposes the rise by tier composition and growth rate. Targeted sub-analyses examine the contributions of COVID-era publication shifts, English-speaking versus non-English-speaking countries, and the post-2022 emergence of large language models. Public trust shapes whether scientific advice is heard and acted upon\cite{wintterlin2022predicting}, and recent surveys document a measurable erosion of public credibility ratings for scientists\cite{annenberg2024credibility}. We complement corpus analysis with a within-subject experiment ($N = 801$; 32{,}040 trials) that provides causal insight into the perceptual consequences of war framing for lay audiences.

\subsection*{Literal military language declines as metaphor accelerates}

\begin{figure}[!ht]
  \centering
  \includegraphics[width=\textwidth]{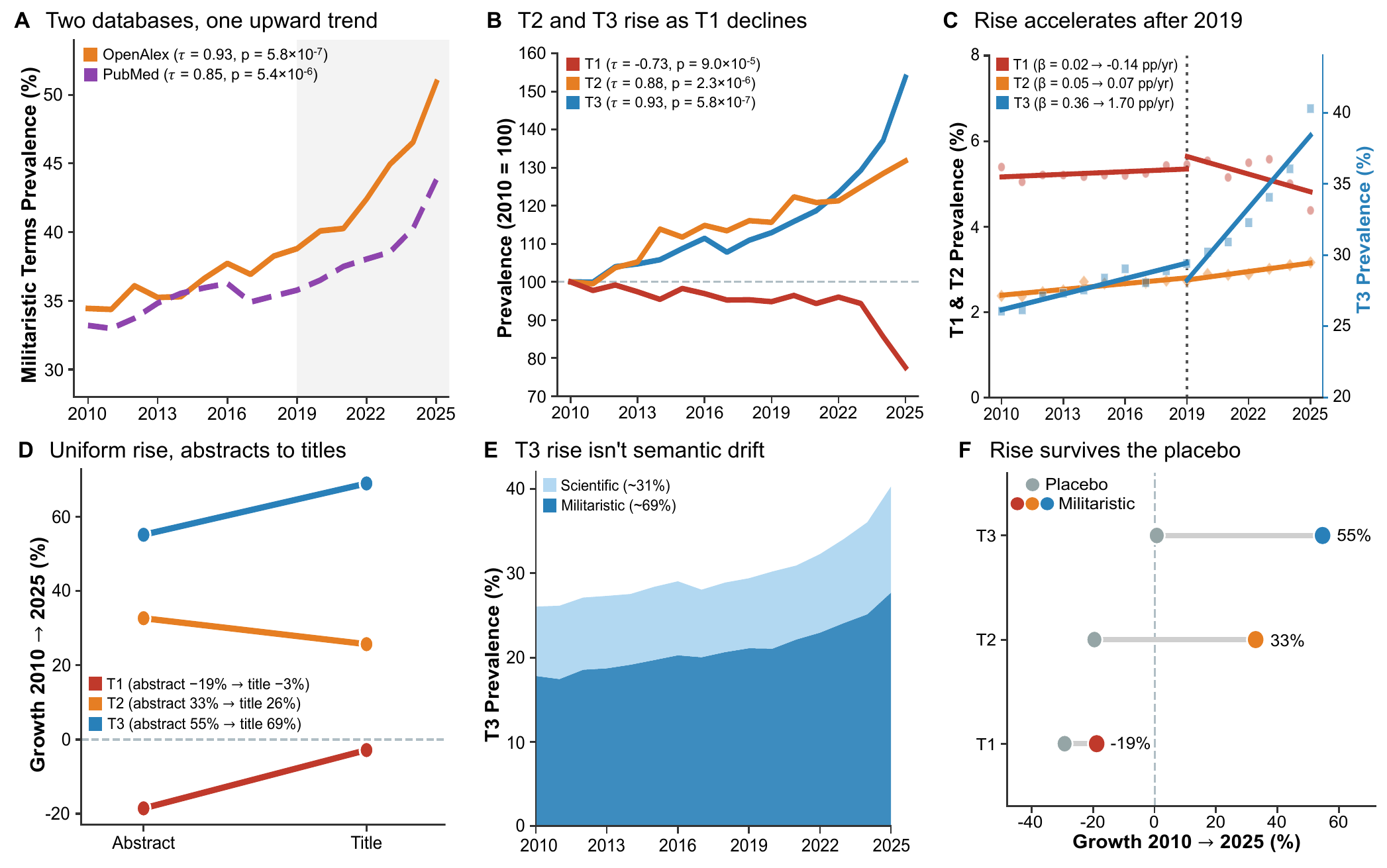}
  \caption{\textbf{Temporal dynamics of use of militarized terms in scientific publications, 2010-2025.}
  \textbf{A}.~Overall prevalence of militaristic terms (all tiers) in OpenAlex (orange) versus PubMed (purple), 2010-2025, with Mann-Kendall trend statistics annotated ($\tau = 0.93$ OpenAlex, $\tau = 0.85$ PubMed). The shaded region marks the post-2019 acceleration.
  \textbf{B}.~Indexed prevalence (2010~$= 100$) for Tier~1 (T1, direct military; crimson), Tier~2 (T2, combat; amber), and Tier~3 (T3, ambiguous; blue), pooled across both databases, with Mann-Kendall $\tau$ annotated per tier.
  \textbf{C}.~Piecewise linear regression with a structural break at 2019 (dotted line) detected by the Pruned Exact Linear Time (PELT) algorithm. Pre- and post-break slopes (percentage points per year) are annotated per tier; T1 and T2 use the left axis, T3 the right axis.
  \textbf{D}.~Relative growth from 2010 to 2025 for each tier, compared between abstracts and titles.
  \textbf{E}.~Large-language-model disambiguation of Tier-3 usage: the share of Tier-3 occurrences classified as militaristic ($\sim$69\%) versus plain-scientific ($\sim$31\%) by an independent classifier (Qwen2.5-3B-Instruct), shown over time.
  \textbf{F}.~Relative growth from 2010 to 2025 of the war lexicon versus a frequency-matched placebo lexicon, by tier.}
  \label{fig:longitudinal}
\end{figure}

The prevalence of militaristic terms in scientific articles rose from 34.4\% in 2010 to 50.9\% in 2025 in OpenAlex (a 47.8\% relative increase) and from 33.2\% to 43.8\% in PubMed (32.0\%; Mann-Kendall $\tau = 0.93$ OpenAlex, $p = 5.8 \times 10^{-7}$; $\tau = 0.85$ PubMed, $p = 5.4 \times 10^{-6}$; Fig.~\ref{fig:longitudinal}a). The fact that these two disjoint corpora, indexed through different curation pipelines, agree on the year-by-year trajectory (Pearson $r = 0.96$, $p < 10^{-8}$; Supplementary Fig.~S2), suggests that the trend is located in scientific writing itself. The tier composition of the rise is the opposite of what literal warfare would predict. Direct military terms (Tier~1; e.g.\ warfare, army, battlefield) declined in titles ($-2.9\%$) and abstracts ($-18.6\%$) over this period. By contrast, metaphorical use (Tier~2; e.g.\ combat, battle, weapon) rose in both (titles $+25.7\%$; abstracts $+32.7\%$), and ambiguous, dual-use terms (Tier~3; e.g.\ attack, defense, target) rose more sharply still (titles $+69.0\%$; abstracts $+55.2\%$; Fig.~\ref{fig:longitudinal}d). Scientists are not increasingly describing literal warfare; they are increasingly importing its structure into other domains.

This pattern is not an artifact of broader lexical drift in scientific writing. A frequency-matched placebo lexicon, sampled to share unigram statistics with the war lexicon, rose by only $-19.6\%$ (Tier~2) and $+0.7\%$ (Tier~3) over the same window, well below the war lexicon's $+32.9\%$ and $+54.7\%$ (Fig.~\ref{fig:longitudinal}f)\cite{burton2021reconsidering}. An independent LLM classifier (Qwen2.5-3B-Instruct) applied to Tier-3 occurrences judged roughly 69\% of dual-use terms to carry a militaristic rather than plain-scientific sense. This majority holds across the study window, indicating the Tier-3 rise reflects genuinely militarized usage rather than neutral vocabulary drift (Fig.~\ref{fig:longitudinal}e; independent LLM tone-scoring corroborates the OpenAlex trend, Supplementary Fig.~S4). When indexed to $2010 = 100$, Tier~1 falls below the baseline while Tier~2 and Tier~3 diverge upward (Fig.~\ref{fig:longitudinal}b). The cross-database trajectory contains a structural inflection at 2019, identified by Pruned Exact Linear Time (PELT) changepoint detection (Fig.~\ref{fig:longitudinal}c). Post-break slopes diverge sharply by tier: Tier-3 ambiguous use accelerates 4.6-fold ($+0.37$ to $+1.70$ percentage points per year), whereas Tier-1 literal use reverses sign (Supplementary Table~S3). This break precedes the COVID-19 pandemic and the post-2022 emergence of LLMs, ruling out either as the sole driver of the observed patterns.

\subsection*{Conflict-involved nations lead, with Global-South catching up}

\begin{figure}[!ht]
  \centering
  \includegraphics[width=\textwidth]{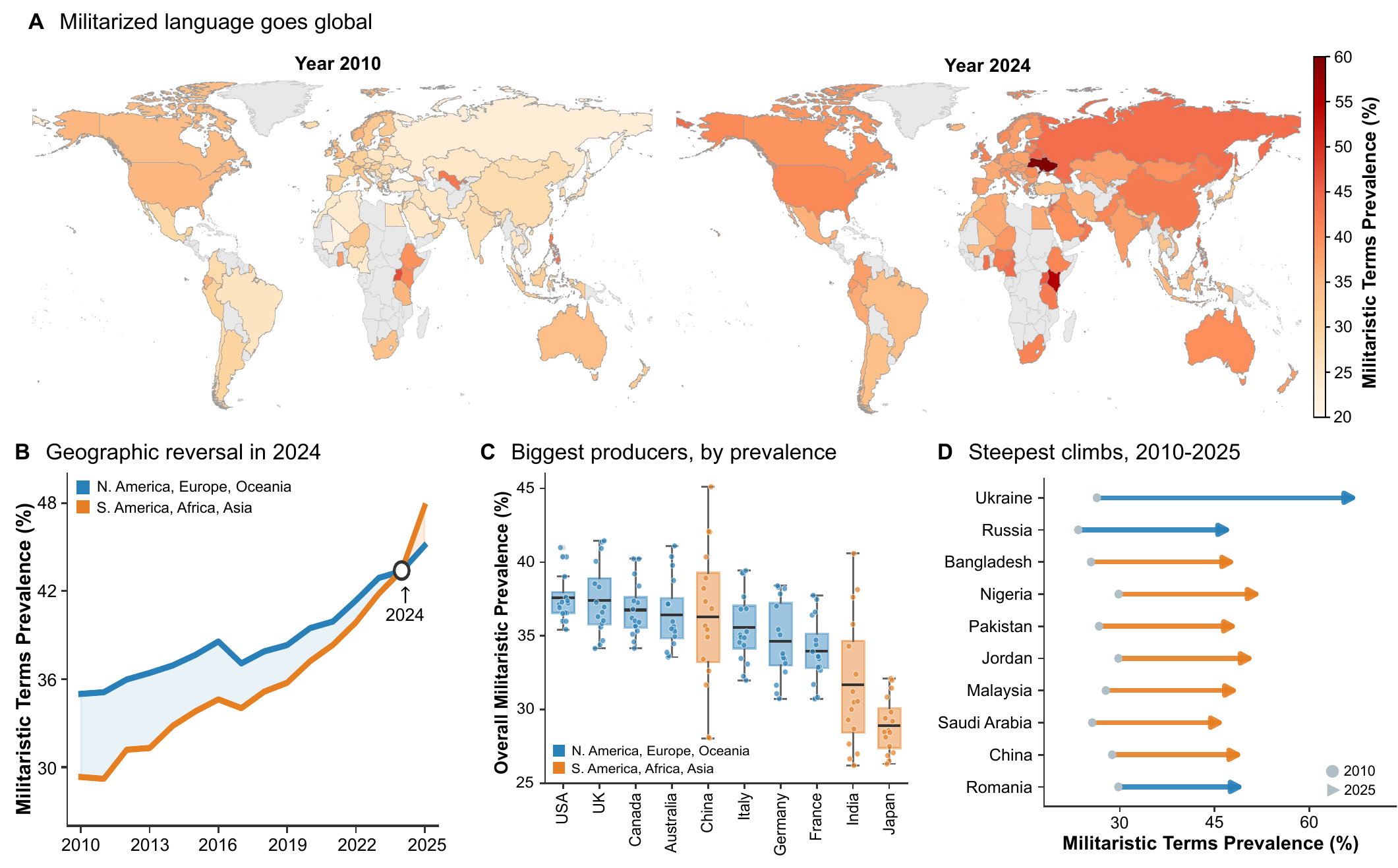}
  \caption{\textbf{Geographic variation in the prevalence of militarized scientific language.}
  \textbf{A}.~Side-by-side world choropleths of the prevalence of militaristic terms by country of author affiliation in 2010 (left) and 2024 (right). Countries with fewer than 10,000 cumulative papers across 2010-2025 are shown in gray.
  \textbf{B}.~Hemispheric trajectories of overall militaristic prevalence, 2010-2025: the Global North (North America, Europe, Oceania) versus the Global South (South America, Asia, Africa), with the 2024 crossover marked.
  \textbf{C}.~Box plots of yearly prevalence for the ten highest-volume countries, sorted by descending mean prevalence. Each point represents one publication year; horizontal lines indicate means. Boxes span the interquartile range.
  \textbf{D}.~Steepest climbs: the ten countries with the largest increase in overall militaristic prevalence from 2010 (filled gray circle) to 2025 (arrow), colored by hemispheric group.}
  \label{fig:geographic}
\end{figure}

The country-level prevalence of militaristic terms varies substantially across the corpus, with conflict-involved nations and East African research systems leading the upper tail. In 2024, country-level prevalence ranges from 31.5\% (Japan) to 71.9\% (Ukraine), with Kenya (54.9\%) second (Fig.~\ref{fig:geographic}a). The cross-country mean rose from 30.5\% in 2010 to 39.6\% in 2024 (Fig.~\ref{fig:geographic}a), and 89 of the 91 countries showed an increase in use of militaristic terms; the number of countries with prevalence at or above 40\% grew sixfold, from 5 to 32. Ukraine alone shows a near-tripling, from 26.3\% in 2010 to 71.9\% in 2024, coincident with the full-scale Russia–Ukraine conflict. The Ukrainian case is a natural experiment in real time: prevalence accelerated sharply after 2022 as warfare entered the country's research environment. The steepest climbs from 2010 to 2025 are led by conflict-involved and Global-South research systems, with Ukraine and Russia rising fastest (Fig.~\ref{fig:geographic}d).

Among the ten highest-volume science producers, prevalence clusters more narrowly, ranging from 28.9\% (Japan) to 37.6\% (United States) (Fig.~\ref{fig:geographic}c). The narrow spread among major producers locates the wider geographic variation in smaller, often conflict-adjacent science systems. The hemispheric pattern reveals a generational reversal across the study period\cite{confraria2017global}. In 2010, the Global North (North America, Europe, Oceania) led with mean prevalence of 35.1\% versus 29.2\% for the Global South (South America, Asia, Africa). By 2025, the Global South had grown by 62.8\% (versus 28.8\% for the North) and overtaken the North at 47.5\% versus 45.2\% (Fig.~\ref{fig:geographic}b). The reversal is concentrated in Tier-3 ambiguous vocabulary, where the Global South rose from 23.7\% to 42.1\% and overtook the Global North's 37.2\%. Tier-3 contains dual-use vocabulary that is increasingly standard across scientific writing; the Global-South catch-up therefore likely reflects converging publication norms more than localized adoption of war framing.

\subsection*{Disciplines converge in prevalence but split on deliberateness}

\begin{figure}[!ht]
  \centering
  \includegraphics[width=\textwidth]{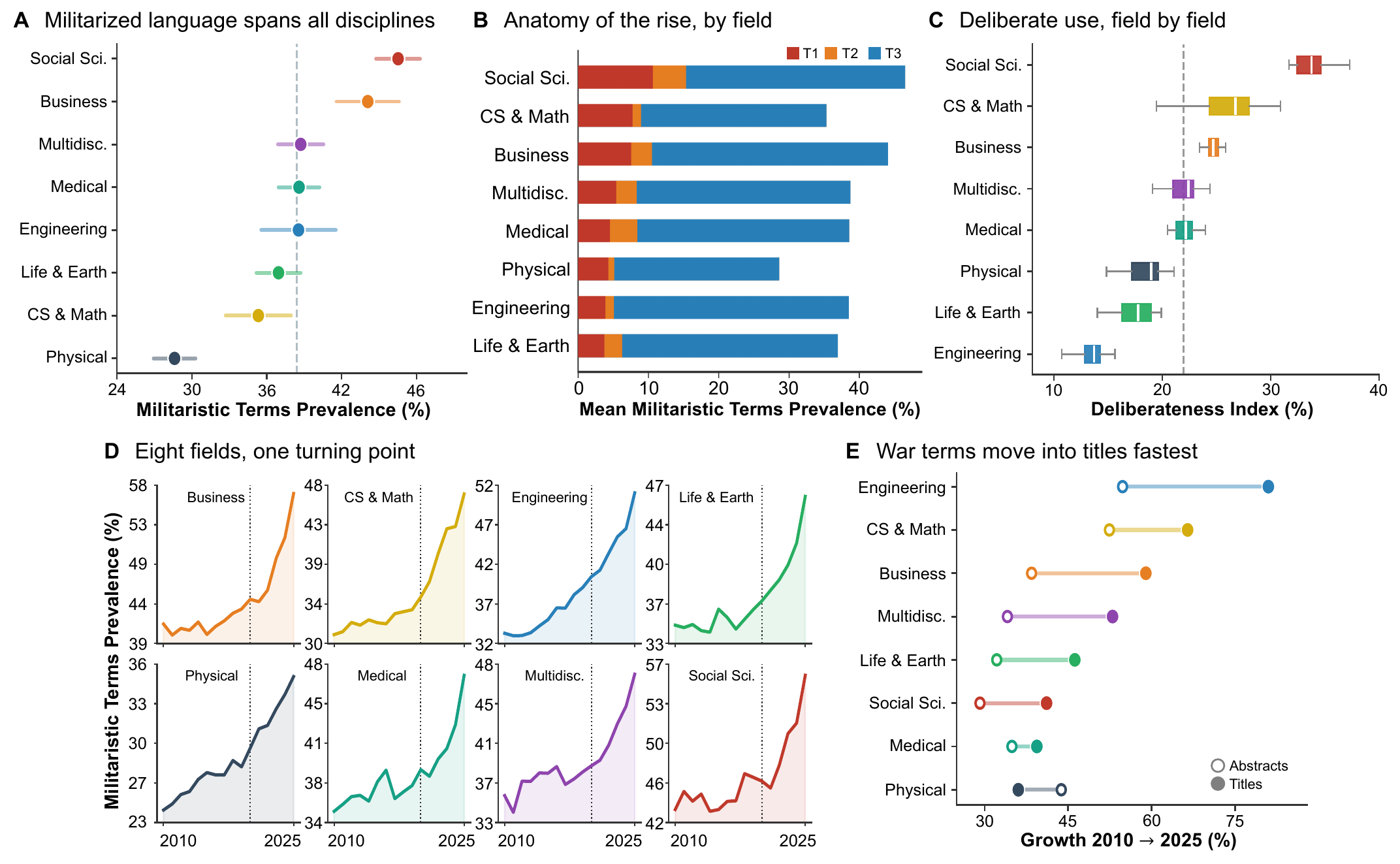}
  \caption{\textbf{Disciplinary patterns of militarized language adoption.}
  \textbf{A}.~Lollipop plot of mean prevalence of militaristic terms by discipline, with horizontal bars showing the range across years (2010-2025).
  \textbf{B}.~Stacked horizontal bars decomposing mean prevalence into T1 (crimson), T2 (amber), and T3 (blue) contributions per discipline. 
  \textbf{C}.~Deliberateness index, defined as (T1\,+\,T2)\,/\,(T1\,+\,T2\,+\,T3)\,$\times$\,100, quantifying the proportion of militaristic language that is unambiguously martial. Box plots show the distribution of yearly values per discipline (median and interquartile range).
  \textbf{D}.~Per-discipline union prevalence over 2010-2025 for each of the eight super-disciplines, shown as filled line plots. The vertical dotted line in each panel marks the 2019 structural break. 
  \textbf{E}.~Growth rates of union prevalence from 2010 to 2025, separately for titles (filled markers) and abstracts (open markers), by discipline.}
  \label{fig:disciplines}
\end{figure}

The disciplinary use of militaristic terms varies substantially in both level and composition. Among the eight super-disciplines, Social Sciences leads at mean overall prevalence of 46.5\%, followed by Business (44.1\%); Physical Sciences trails at 28.6\% (Fig.~\ref{fig:disciplines}a). Tier composition by discipline reveals two distinct patterns of usage. Social Sciences shows the highest Tier-1 (literal military) share at 10.6\%, the largest deliberate use among disciplines. Engineering shows the inverse pattern: low Tier-1 (3.9\%) and high Tier-3 (33.5\%), pointing to incidental, dual-use vocabulary (Fig.~\ref{fig:disciplines}b). The Deliberateness Index, defined as $(T_1 + T_2)/(T_1 + T_2 + T_3) \times 100$, makes this contrast explicit. Social Sciences ranks highest at 33\% and Engineering lowest at 13\% (Fig.~\ref{fig:disciplines}c).

The growth rates over 2010-2025 broadly invert the disciplinary ranking, with low-level fields growing fastest. Engineering and Computer Science grow fastest at $+54.9\%$ and $+51.6\%$, respectively; Social Sciences grows slowest at $+27.5\%$ (Fig.~\ref{fig:disciplines}e). The inverse pattern is consistent with broad lexical convergence\cite{vilhena2014finding} rather than localized adoption of war framing. The per-discipline changepoint analysis identifies a coordinated 2019 PELT break in all eight super-disciplines (Fig.~\ref{fig:disciplines}d). Disciplinary heterogeneity therefore concerns level and tier composition; the timing of the rise is shared.

\subsection*{Prevalence tracks armed conflict at multiple scales}

\begin{figure}[!ht]
  \centering
  \includegraphics[width=\textwidth]{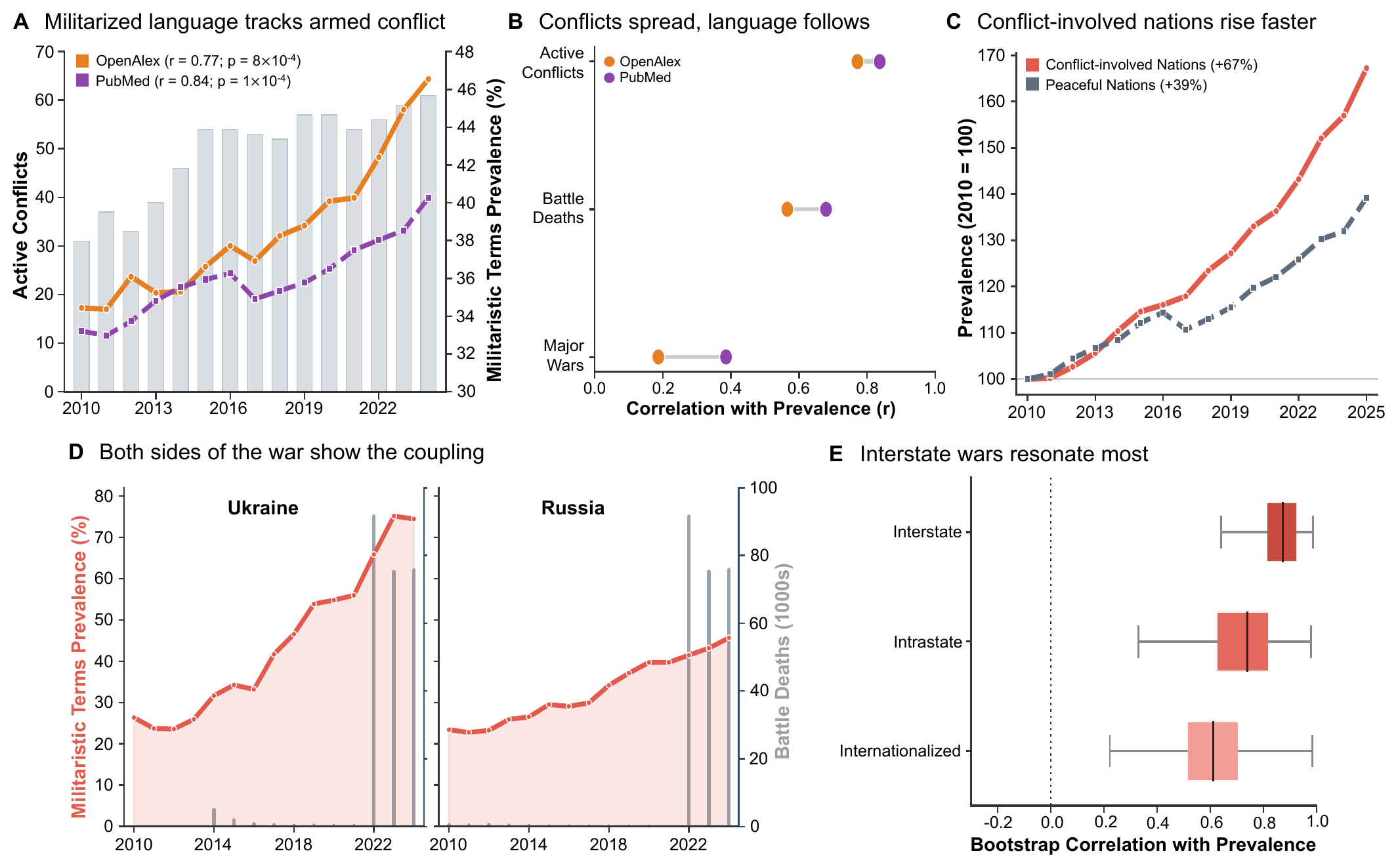}
  \caption{\textbf{Association between armed conflict and militarized scientific language.}
  \textbf{A}.~Global overlay of the number of active armed conflicts per year (UCDP/PRIO, gray bars; left axis) and overall militaristic-term prevalence (right axis) in OpenAlex (orange) and PubMed (purple), with Pearson correlations annotated.
  \textbf{B}.~Correlation of overall prevalence with three conflict indicators: active-conflict counts, battle-death totals, and major-war counts, shown separately for OpenAlex (orange) and PubMed (purple).
  \textbf{C}.~Indexed prevalence (2010\,$=$\,100) for conflict-involved countries (red; $+67\%$) versus peaceful countries (dark gray dashed; $+39\%$).
  \textbf{D}.~Case studies of Ukraine and Russia. Lines: union prevalence (left axis); bars: UCDP battle deaths (right axis). Both belligerents show prevalence rising with conflict intensity across the 2014 and 2022 escalations.
  \textbf{E}.~Bootstrap correlation between prevalence and conflict activity by conflict type (interstate, intrastate, internationalized). Boxes show the bootstrap distribution; the dotted line marks zero correlation.}
  \label{fig:conflict}
\end{figure}

Three lines of evidence link the rise to armed conflict: annual correlation with global conflict counts, elevated prevalence in conflict-involved countries, and step-changes around individual conflict events. The annual prevalence and active UCDP/PRIO conflict counts are strongly correlated in both databases (Pearson $r = 0.77$, $p = 8 \times 10^{-4}$ in OpenAlex; $r = 0.84$, $p = 1 \times 10^{-4}$ in PubMed; Fig.~\ref{fig:conflict}a). The association is specific to the intensity measure: across candidate conflict indicators, prevalence tracks the annual count of active armed conflicts most closely, with weaker coupling to global battle-death totals and weakest to the count of major wars, in both databases (Fig.~\ref{fig:conflict}b). The number of ongoing conflicts, rather than their lethality, is the dimension that scientific language most closely follows.

When indexed to 2010, conflict-involved countries grew faster than peaceful countries over the study period ($+67\%$ versus $+39\%$; Fig.~\ref{fig:conflict}c). This gap widens after 2019, when conflict-involved nations accelerate more steeply than peaceful ones (Supplementary Fig.~S3). Two case studies from opposite sides of the same war sharpen the link: both Ukraine and Russia show prevalence rising in step with battle-death intensity across the 2014 and 2022 escalations, indicating the coupling is not confined to the country under attack (Fig.~\ref{fig:conflict}d). When disaggregated by conflict type, the coupling is strongest for interstate wars, weaker for intrastate conflicts, and weakest for internationalized ones (Fig.~\ref{fig:conflict}e). The annual, country-level, and case-study evidence converge on the same conflict-aligned pattern.

\subsection*{The COVID and LLM eras accelerated the rise across disciplines and language groups}

\begin{figure}[!ht]
  \centering
  \includegraphics[width=\textwidth]{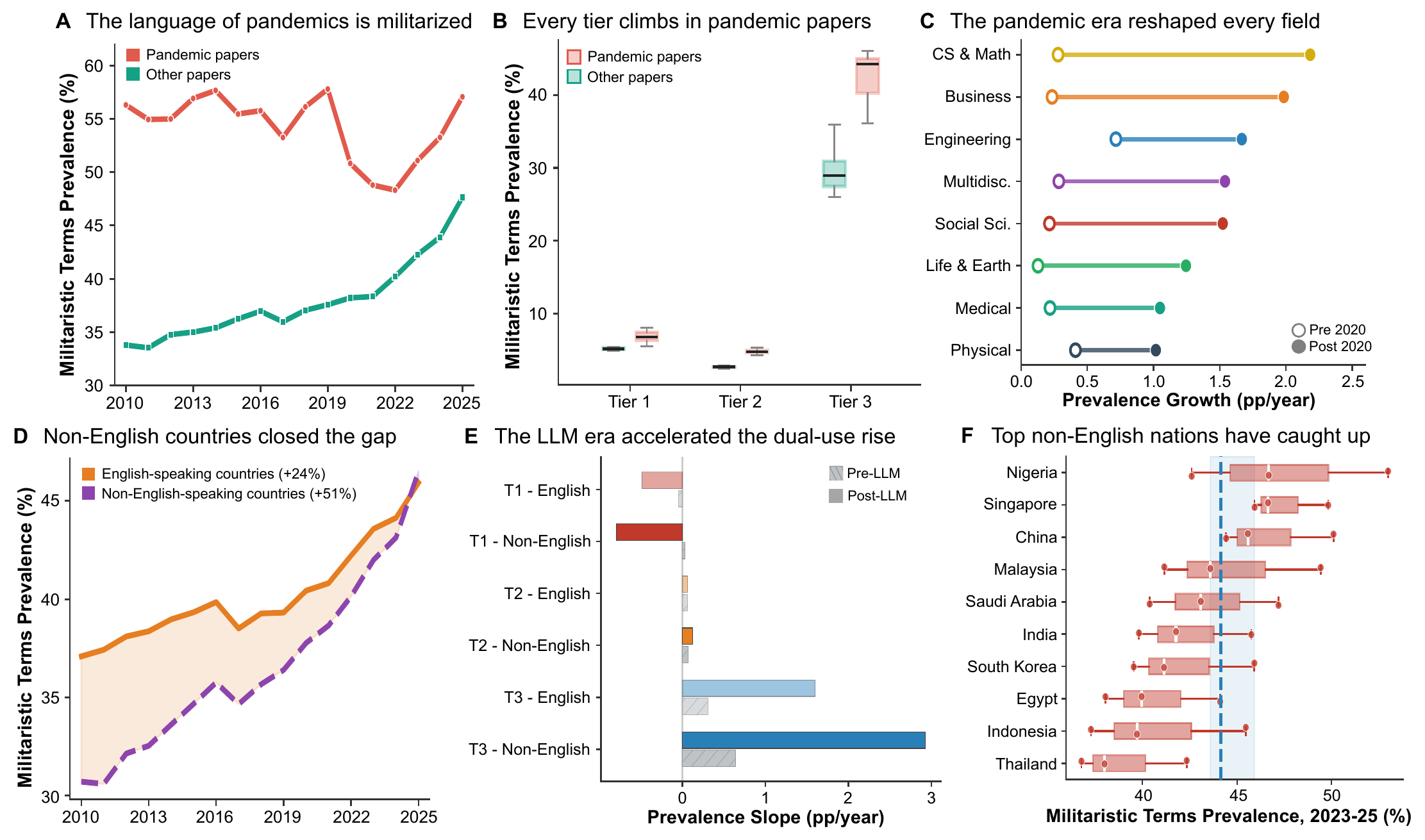}
  \caption{\textbf{The COVID-19 pandemic and the LLM era accelerated militarized language across disciplines and language groups.}
  \textbf{A}.~Pandemic-related papers (red) versus all other papers (teal): overall militaristic-term prevalence, 2010-2025.
  \textbf{B}.~Tier-specific prevalence in pandemic-related versus other papers. Box plots show the distribution of yearly values; every tier is elevated in pandemic papers.
  \textbf{C}.~Per-discipline prevalence growth (percentage points per year) before 2020 (open markers) versus after 2020 (filled markers), for each of the eight super-disciplines.
  \textbf{D}.~Overall prevalence for English-speaking countries (orange; $+24\%$) versus non-English-speaking countries (purple; $+51\%$), 2010-2025.
  \textbf{E}.~The LLM era and the dual-use rise. Pre-LLM (2010-2022, hatched) versus post-LLM (2023-2025, filled) prevalence slopes (percentage points per year) for each tier, stratified by language group.
  \textbf{F}.~Top ten non-English-speaking countries by militaristic-term prevalence in the post-LLM period (2023-2025). Box plots show yearly values; the blue dashed line and shading mark the native-English-speaking median and range.}
  \label{fig:sensitivity}
\end{figure}

The pandemic era did not merely coincide with the rise; it amplified it. Pandemic-related papers carry higher militaristic prevalence than non-pandemic papers across the entire study window, with the overall gap largest in the 2020-2024 mean (50.0\% versus 40.8\%; Fig.~\ref{fig:sensitivity}a). This elevation holds within every tier: pandemic papers show higher Tier-1, Tier-2, and Tier-3 prevalence, with metaphorical Tier-2 use elevated by a factor of 1.80 (Fig.~\ref{fig:sensitivity}b). The tendency of pandemic-themed scientific writing to import militaristic structure pre-dates COVID-19 and intensified during it, consistent with prior reports of war framing in pandemic communication\cite{wicke2020framing,semino2021fire}. In a comparison of pre-2020 versus post-2020 prevalence growth across disciplines, the acceleration appears in all eight super-disciplines, not only pandemic-adjacent biomedical fields (Fig.~\ref{fig:sensitivity}c). The COVID-era amplification is broad-based and structural, spanning the full disciplinary range.

The post-2022 LLM transition reshaped the language stratification of the rise. Native-English-speaking countries led throughout most of the study window, but non-English-speaking countries grew far faster ($+51\%$ versus $+24\%$) and closed the gap by 2025 (Fig.~\ref{fig:sensitivity}d). The non-English-speaking countries accelerate Tier-2 metaphorical use 1.8-fold and Tier-3 ambiguous use 4.6-fold across the LLM transition, while English-speaking countries show 1.0-fold for Tier-2 and 5.1-fold for Tier-3 (Fig.~\ref{fig:sensitivity}e). The Tier-3 acceleration is comparable across language groups, but the Tier-2 metaphorical jump is concentrated in non-English-speaking countries, consistent with reports that LLM-mediated writing tends to converge non-native English prose toward higher-register lexical patterns\cite{liang2025quantifying,kobak2025delving}. Among the ten highest-volume non-English-speaking countries (each with at least 100{,}000 papers), nine crossed the native-English-speaking median (44.1\%) in 2025, with only Thailand remaining below (Fig.~\ref{fig:sensitivity}f). Across both the COVID and LLM eras, the rise broadened and became more linguistically uniform, extending across disciplines and narrowing the gap between native-English-speaking countries and countries where English is not the native language.

\subsection*{War framing depresses credibility, with correlated funding-willingness loss}

\begin{figure}[!ht]
  \centering
  \includegraphics[width=\textwidth]{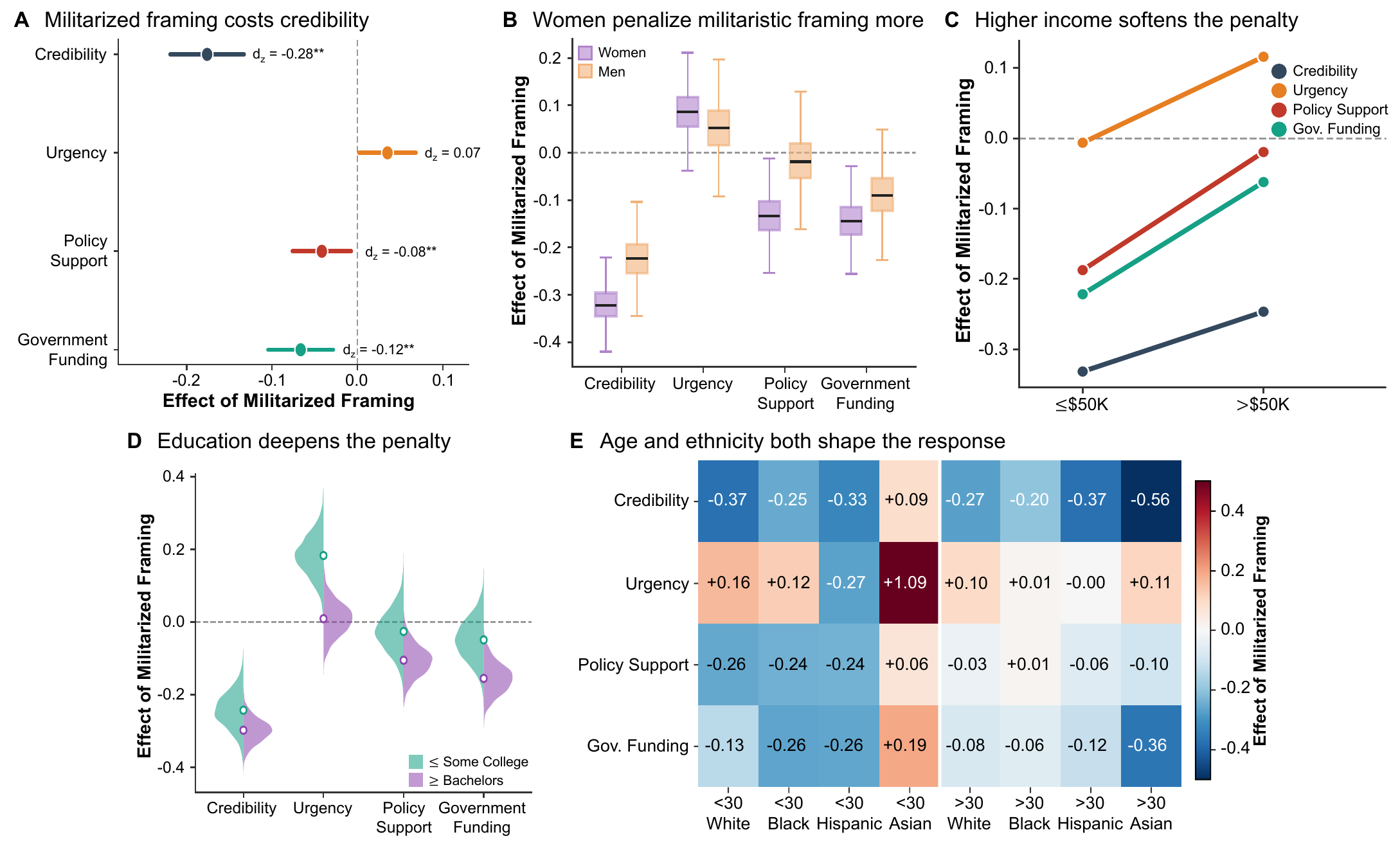}
  \caption{\textbf{Survey experiment: effect of militarized framing on public perceptions of science.}
  \textbf{A}.~Forest plot of the within-subject effect of war-framed versus neutrally-framed scientific abstracts on four outcome measures (7-point Likert scales). Points show mean differences; horizontal lines show 95\% confidence intervals; Cohen's $d_z$ is annotated per outcome. $^{**}p < 0.05$.
  \textbf{B}.~Gender moderation. Per-subject box plots of the framing effect for women (purple) versus men (orange) across the four outcomes.
  \textbf{C}.~Income moderation. Framing effect for respondents with household income $\leq$\$50K versus $>$\$50K, with lines connecting the two income groups for each outcome (colored by outcome).
  \textbf{D}.~Education moderation. Split-violin distributions of the framing effect for respondents with $\leq$~some college (teal) versus $\geq$~a Bachelor's degree (purple), by outcome; markers indicate point estimates.
  \textbf{E}.~Intersectional heatmap of Cohen's $d_z$ across age ($<$30, $\geq$30) and race (White, Black, Hispanic, Asian) subgroups for each outcome.}
  \label{fig:survey}
\end{figure}

In a US convenience sample of 801 adults recruited via the Prolific online research platform, within-subject exposure to war-framed versus neutrally-framed scientific abstracts produced an asymmetric pattern of effects across four outcomes of interest (Fig.~S1; sample construction and design-balance checks in Supplementary Table~S5). War framing reduced perceived credibility by 0.176 Likert units (95\% CI $[-0.21, -0.14]$; $d_z = -0.28$, $p < 10^{-20}$), willingness to let government fund the work by 0.066 units (CI $[-0.11, -0.02]$; $d_z = -0.12$, $p = 0.002$), and policy support by 0.042 units (CI $[-0.08, -0.001]$; $d_z = -0.08$, $p = 0.044$). The perceived urgency rose modestly by 0.035 units (CI $[-0.005, +0.075]$; $d_z = +0.07$, $p = 0.082$; Fig.~\ref{fig:survey}a). The credibility effect is roughly twice the magnitude of the next-largest outcome and is detectable in every demographic subgroup tested. Within both women ($n = 456$) and men ($n = 329$), war framing significantly reduced perceived credibility ($d_z = -0.32$ women, $d_z = -0.22$ men; both $p < 0.001$), with within-women effects on funding ($d_z = -0.14$, $p = 0.002$) and policy support ($d_z = -0.13$, $p = 0.005$) also reaching significance while the corresponding within-men effects on funding and policy did not. The formal gender $\times$ condition interaction was not significant for any outcome (credibility $\beta = -0.06$, 95\% CI $[-0.14, +0.01]$, $p = 0.09$; Fig.~\ref{fig:survey}b). Income moderation is sharper: lower-income respondents ($\leq$\$50K, $n = 261$) show the largest credibility, policy, and funding decrements ($d_z = -0.33$, $-0.19$, $-0.22$, all $p < 0.01$) and no urgency response, whereas higher-income respondents ($>$\$50K, $n = 514$) show a smaller credibility decrement ($d_z = -0.25$) and are the only subgroup with a significant urgency increase ($d_z = +0.12$, $p = 0.005$); the income $\times$ condition interaction is significant for credibility ($\beta = +0.10$, $p = 0.015$), policy ($\beta = +0.09$, $p = 0.033$), and funding ($\beta = +0.11$, $p = 0.022$; Fig.~\ref{fig:survey}c). Education stratification reveals divergent response channels by outcome: respondents with at most some college ($n = 306$) show a significant urgency increase ($d_z = +0.18$, $p = 0.001$) but no funding or policy response, whereas respondents with a Bachelor's degree or higher ($n = 491$) show significant funding ($d_z = -0.15$, $p < 0.001$) and policy ($d_z = -0.11$, $p = 0.017$) attenuation but no urgency increase; both groups show comparable credibility decrements ($d_z = -0.24$ and $-0.30$, both $p < 0.001$). The war $\times$ education interaction is significant for credibility ($\beta = -0.08$, $p = 0.047$) but not for urgency ($\beta = -0.08$, $p = 0.052$) (Fig.~\ref{fig:survey}d). The credibility attenuation is universal across gender, income, and education, but the downstream funding and policy consequences are concentrated in women, lower-income, and higher-education respondents. The credibility penalty likewise holds across all five scientific domains tested, from immunology to ecology (Supplementary Fig.~S5).

Intersectional stratification by age and race confirms the credibility attenuation across most cells. The Age $\times$ Race heatmap of Cohen's $d_z$ shows credibility decrements in seven of the eight age-race subgroups, ranging from $d_z = -0.20$ in older Black respondents ($n = 147$) to $d_z = -0.56$ in older Asian respondents ($n = 31$); the largest effects fall in older Asian, older Hispanic ($d_z = -0.37$, $n = 48$) and younger White ($d_z = -0.37$, $n = 78$) cells, while the single positive cell (younger Asian respondents, $n = 9$) is too small for interpretation (Fig.~\ref{fig:survey}e). At the individual level, per-respondent framing-shifts in credibility and funding willingness are correlated across people (Pearson $r = 0.63$, $p = 8.2 \times 10^{-91}$; the four outcomes form coherent assessment and action constructs, Supplementary Table~S4), and within-person credibility and funding ratings are coupled trial-by-trial (repeated-measures $r = 0.66$, 95\% CI $[0.66, 0.67]$; crossed random-effects multilevel $\beta = 0.76$, 95\% CI $[0.75, 0.77]$, $p < 10^{-15}$; 32{,}040 trials in 801 respondents). Respondents who downgraded credibility under war framing also downgraded their willingness to let government fund the work, placing the credibility-funding link at the level of individual perception rather than emerging only in group means. The credibility hit therefore has a measurable downstream correlate in funding intentions, completing a chain from rising war vocabulary in scientific writing through reader perception to a behavioral intention.

\section*{Discussion}

Scientific writing has acquired the vocabulary of war, even as scientists describe less literal warfare. Across 21.4 million peer-reviewed articles from OpenAlex and PubMed (2010-2025), the prevalence of militaristic terms rose by 48\% and 32\%, respectively, with two disjoint corpora, indexed through different curation pipelines, agreeing on the year-by-year trajectory ($r = 0.96$). The rise is concentrated in metaphorical and ambiguous, dual-use vocabulary; direct military reference declined. The country-level prevalence is conflict-aligned, with Ukraine and East African research systems leading the upper tail. The per-discipline pattern accelerates uniformly at a 2019 inflection point shared across all eight super-disciplines. The post-2022 emergence of LLMs then accelerated the rise and narrowed the gap between native-English-speaking and non-English-speaking countries. Despite this rise in militaristic language in scientific works, and in line with concerns that war metaphors may undermine credibility\cite{hauser2015war, hendricks2018emotional, thibodeau2011metaphors}, a within-subject experiment ($N = 801$; 32{,}040 trials) shows that war framing in scientific writing depresses credibility among a lay-audience by $d_z = -0.28$, with the loss correlated within individuals to a decrement in willingness-to-fund ($r = 0.63$). The vocabulary that scientists choose carries measurable downstream costs in how their work is received.

The mechanism of metaphorical adoption lies in the composition of the rise, not its volume. Direct military vocabulary declined while metaphorical and ambiguous use accelerated, indicating that scientists are importing the structure of war into other domains rather than describing it. The acceleration is universal across 89 of 91 countries, all eight super-disciplines, and the post-LLM convergence between native-English-speaking and non-English-speaking countries, suggesting a shared discursive substrate rather than localized adoption of war framing. At the reader level, within-person credibility and funding ratings are tightly coupled on a trial-by-trial basis (repeated-measures $r = 0.66$, 95\% CI $[0.66, 0.67]$; crossed random-effects multilevel $\beta = 0.76$, 95\% CI $[0.75, 0.77]$, $p < 10^{-15}$; 32{,}040 trials nested in 801 respondents), placing the credibility-funding link at the level of individual perception rather than group means. Respondents who downgraded credibility under war framing also downgraded their willingness to let government fund the work.

Our findings extend a literature that has been fragmented across disciplines. The 1978 critique of war metaphors in cancer medicine\cite{sontag1978illness} predicted moral burdens on patients. We confirm at the population scale that the vocabulary it warned against has spread far beyond oncology, accelerating across all eight super-disciplines and converging across language groups. Experimental work has documented credibility costs, reduced engagement and emotional consequences of individual war frames\cite{hauser2015war,hendricks2018emotional,thibodeau2011metaphors}; our within-subject experiment confirms the credibility cost ($d_z = -0.28$) in scientific writing specifically. The methodological caution that lexicon-prevalence findings can be artifacts of generic text-statistics drift\cite{burton2021reconsidering} is met by our frequency-matched placebo lexicon, which showed no comparable rise while the war lexicon rose sharply. The science-of-science tradition has tracked topical drift, citation patterns and authorship dynamics\cite{foster2015tradition,liu2023data} but rarely the rhetorical register of scientific writing itself; we add this dimension. Recent population-scale lexical work has focused on LLM-marker vocabulary\cite{liang2025quantifying,kobak2025delving}; our analysis reveals a parallel but distinct trajectory in war vocabulary that interacts with the LLM transition but predates it.

There are four limitations that bound the inferences this analysis can support. First, our corpus analysis screens titles and abstracts only, not full body text; prevalence in the most-read parts of papers therefore captures framing density at the level of how work is presented to other scientists and to indexed readers, but misses methods and discussion sections where war metaphors may also commonly arise. Second, lexicon-based prevalence measures word use, not intent: a hand-curated three-tier lexicon cannot distinguish whether an author deliberately chose "combat" over "experiment" or absorbed it as conventional vocabulary. Third, our reader survey is single-session and drawn from US Prolific respondents; we cannot speak to cross-cultural reception of war framing or to longitudinal effects of repeated exposure. Fourth, corpus prevalence and individual reader perception do not directly measure downstream uptake in news media, policy discourse, or public conversation, where the vocabulary's effects on lay reasoning would ultimately compound.

There are two main lines of future work that would further clarify the mechanism and extend the scope of this effort. Interview studies of scientists could distinguish deliberate war framing from conventional absorption, addressing the intent-versus-language gap. A cross-cultural extension of the reader experiment beyond US English audiences would test whether the credibility cost generalizes or amplifies in different rhetorical traditions. The normative implication is direct: the words scientists choose are not stylistic flourishes but cognitive primers, and the cumulative drift toward war framing in science has measurable costs in the lay perception of legitimacy. What we describe as war begins to feel like one.

\section*{Methods}

\subsection*{Data sources and preprocessing}

The corpus was assembled from two complementary bibliographic databases: OpenAlex (multidisciplinary)\cite{priem2022openalex} and PubMed (biomedical). The two databases were chosen because their indexing pipelines are distinct (OpenAlex builds from open-access metadata sources, while PubMed is curated by the US National Library of Medicine), so cross-database concordance functions as a corpus-independent replication check rather than a within-source consistency test. For each database, we downloaded paper records covering the period January 2010 through August 2025, retaining the following metadata per record: title, abstract, publication year, journal name, document type and author affiliations. Both databases were queried day-by-day for each publication date in this period. OpenAlex records were retrieved via the OpenAlex API, with records carrying a PubMed ID excluded at download time to avoid overlap with the PubMed corpus. PubMed records were retrieved via the NCBI E-utilities API, which caps results at 10{,}000 records per query; days on which more than 10{,}000 papers were indexed were truncated to the first 10{,}000 records returned. OpenAlex and PubMed are treated effectively as independent samples throughout. Records were filtered to keep only items with abstracts of at least 20 words and non-empty titles; no further document-type filter was applied, so editorials, letters and reviews were retained alongside research articles. The title and abstract text were cleaned by lowercasing, replacing punctuation with spaces and collapsing repeated whitespace. After applying abstract and title filters, $N = 21{,}404{,}544$ papers (OpenAlex: 11{,}876{,}194; PubMed: 9{,}528{,}350) were retained for downstream analysis.

\subsection*{Country attribution}

The per-paper country attribution was carried out in two stages. We first extracted all unique affiliation substrings from the corpus by splitting each paper's multi-author affiliation field into individual affiliations and pooling these across both databases. Each unique substring was then passed through a hand-curated rule cascade that assigned a country in the following order: (i) suffix matching against a country-name lookup covering more than 200 entries, including alternate forms (e.g., P.R. China, Russian Federation), common abbreviations (USA, UK, UAE, KSA) and French, Spanish, German, Portuguese, Turkish, Cyrillic and CJK script variants; (ii) US state-abbreviations patterns at the end of the string (e.g., CA 90210 USA); (iii) full names of US states, Canadian provinces, Indian states, Malaysian states and Indonesian and Nigerian regional locations; (iv) Chinese province and major-city recognition; (v) full-text search for any country name of four or more characters appearing anywhere in the affiliation; (vi) major-city fallback against a curated city-to-country list; and (vii) institution-keyword matching for research bodies whose location is unambiguous (INSERM $\rightarrow$ France, Max Planck $\rightarrow$ Germany, IIT $\rightarrow$ India, Karolinska $\rightarrow$ Sweden and similar). Affiliations failing all rules were marked Unknown and excluded from country-level analyses. The rule cascade was iteratively refined against six independent 500{,}000-record samples drawn from approximately 29.5 million unique affiliation substrings. The countries with at least 10{,}000 papers across the study window were retained, yielding 91 nations for country-level analysis.

Each paper was then attributed to each country detected in its affiliation field: a paper with a US and a UK author contributes one paper-count to both countries. The country-level prevalence in this paper is therefore the joint signal of all authoring countries, not a first-author-only signal.

\subsection*{Discipline attribution}

The per-paper disciplinary attribution was derived from the journal in which each paper was published, with journals mapped to one of eight super-disciplines: Medical \& Health Sciences, Engineering \& Technology, Computer Science \& Mathematics, Physical Sciences, Life \& Earth Sciences, Social Sciences (including Arts \& Humanities), Business \& Management and Multidisciplinary \& General Science. The mapping was constructed in three stages: an initial rule-based pass, a large language model correction pass, and a final precision-rule refinement.

The initial rule-based pass matched journal names against category-specific keyword lists using word-boundary regex (so the keyword cardiology would match Journal of Cardiology but not Acta Endocrinologica). The keyword lists included English, German, Spanish, French, Indonesian, Chinese and other-language variants; first match wins across the eight categories, with the multidisciplinary label as the default for non-matching journals. The first-pass categorization was then passed to a large language model (claude-sonnet-4-5; Anthropic) for review and correction in batches of 1{,}000 (journal, current category) pairs. The system prompt constrained outputs to the same eight-label vocabulary, named the multidisciplinary label as a last-resort fall-back and supplied specific routing rules for ambiguous cases (e.g., medical congresses $\rightarrow$ Medical \& Health Sciences; pure mathematics and statistics $\rightarrow$ Computer Science \& Mathematics; humanities, education, law $\rightarrow$ Social Sciences (including Arts \& Humanities)). The LLM-corrected file was then passed through a final rule-based refinement using more precise word-boundary matching with explicit phrase-versus-word-token distinction, applied in a fixed priority order (Medical \& Health Sciences first, Multidisciplinary as final fallback) to disambiguate edge cases. The resulting journal-to-category map was joined to every paper in the corpus by exact journal-name match.

\subsection*{Lexicon construction}

Militaristic terms were screened using a three-tier hand-curated lexicon designed to separate literal military reference (Tier~1) from metaphorical scientific use (Tier~2) and dual-use, ambiguous vocabulary (Tier~3). The initial candidate list was assembled by the authors through structured brainstorming, drawing on prior war-metaphor literature \cite{sontag1978illness,lakoff1980metaphors,flusberg2017metaphors}. The tier assignments were made against a precision criterion (Tier~1: dominant usage is military; Tier~2: strong military denotation but documented figurative scientific use; Tier~3: non-military scientific use comparably or more common) and reviewed in author consultation. The borderline terms were resolved by the precision criterion and inspection of representative scientific abstracts. Tier~1 contains 51 unambiguously military base terms (e.g., warfare, army, navy, battlefield, combatant, infantry, artillery, terrorism, guerrilla, kamikaze, homeland security, weapons of mass destruction). Tier~2 contains 46 moderately martial terms whose figurative use within scientific discourse is interpretable but not fully ambiguous (e.g., combat, battle, weapon, invasion, enemy, adversary, retreat, casualty, escalation, hostility, sabotage, mobilize, cyberwarfare). Tier~3 contains 44 high-frequency, dual-use terms whose military reading is context-dependent (e.g., attack, defense, target, strategy, campaign, conflict, threat, victory, security, force, mission, operation). The single-word term "intelligence" was assigned to Tier~3 rather than Tier~2 because its dominant scientific usage (artificial intelligence, cognitive intelligence) is non-military.

Each tier was checked separately on the cleaned title and abstract via word-boundary regex matching. Each base term was automatically expanded to admit a regular trailing plural (so weapon matches both weapon and weapons), and explicit irregular plural forms (e.g., armies, enemies, casualties, strategies) were listed separately so they were matched without grammatical inflection. The per-paper title and abstract counts were summed to give a per-paper count for each tier.

The tier-specific prevalence is the percentage of papers containing at least one term from that tier; we use binary per-paper presence rather than per-paper count so that a paper with five Tier-2 terms contributes the same to prevalence as a paper with one, preventing term-dense reviews from dominating the metric. The overall prevalence reported throughout this paper is the sum of the three tier-specific prevalences.

\subsection*{LLM-based Tier-3 disambiguation and tone scoring}

Tier~3 terms are dual-use by construction, so a Tier~3 keyword match does not guarantee a militaristic reading. We therefore passed each Tier~3 occurrence through a contextual disambiguation step, using a small open-weight instruction-tuned language model (Qwen2.5-3B-Instruct), to quantify how often the matches carry a genuinely military sense. For every Tier~3 hit we extracted a symmetric context window of 30 words on each side of the matched term and prompted the model to decide whether the term was used as a war or military metaphor or as a literal, technical scientific term, constraining the reply to a single label (Metaphor or Literal). The decoding was greedy (temperature $0$) for reproducibility. The per-occurrence labels, aggregated across the corpus, yield the militaristic-versus-scientific split reported for Tier~3 (Fig.~\ref{fig:longitudinal}e); this classifier is used only to characterize the composition of Tier~3 usage and does not alter the keyword-based prevalence series.

As an independent check on the lexicon-based measure, we additionally scored the overall militaristic tone of each abstract with the same model. Each abstract was rated on an integer scale from $0$ (completely neutral, standard scientific language) to $5$ (heavily militarized language permeating the text), following a fixed rubric supplied in the prompt, again with greedy decoding and the response constrained to the numeric score. The mean annual tone was then compared with the keyword-based prevalence trend as a convergence validity analysis (Supplementary Fig.~S4). The tone metric is a corroborating measure computed independently of the tier lexicon and is not combined with the prevalence counts.

\subsection*{Placebo lexicon}

A frequency-matched placebo lexicon was constructed to distinguish substantive change in militaristic terms from generic lexical drift across scientific writing\cite{burton2021reconsidering}. The placebo lexicon contains three tiers chosen to share unigram statistics roughly with T1, T2 and T3. P1 (Instrumentation) contains 20 rare, unambiguous lab technique and method terms with frequencies comparable to T1 (e.g., microscopy, chromatography, tomography, spectrometry, calorimetry). P2 (Process) contains 19 medium-frequency scientific-process terms with frequencies comparable to T2 (e.g., calibration, normalization, quantification, polymerization, fractionation). P3 (Reasoning) contains 20 common analytical and statistical terms with frequencies comparable to T3 (e.g., hypothesis, measurement, coefficient, variance, gradient, threshold).

The placebo lexicon was applied to the same corpus using identical word-boundary regex matching with optional plural suffixes and aggregated into yearly placebo prevalence series in the same format as the war-lexicon output. The trends in the placebo lexicon serve as a reference for the expected magnitude of change attributable to general lexical drift in scientific writing: where the war lexicon rises substantially faster than its frequency-matched placebo counterpart, the difference cannot be explained by generic vocabulary drift across the corpus.

\subsection*{Statistical analyses}

\subsubsection*{Trend tests}
The Mann-Kendall non-parametric trend tests were applied to each tier's yearly prevalence series, separately for OpenAlex, PubMed, and the pooled cross-database series. We report Kendall's $\tau$, the Sen slope, and the trend $p$-value for each test.

\subsubsection*{Changepoint detection}
To identify structural breakpoints in the cross-database trajectory and per-discipline series, we employed the Pruned Exact Linear Time (PELT) algorithm\cite{killick2012optimal}. This approach utilized a radial basis function (RBF) kernel, with adaptively selected penalty parameters for each individual series.

\subsubsection*{Cross-database concordance}
We assessed the concordance between the annual prevalence of OpenAlex and PubMed using Pearson and Spearman rank correlations applied to year-paired data.

\subsubsection*{Conflict correlation}
The annual prevalence was correlated with the count of active armed conflicts per year derived from the UCDP/PRIO Armed Conflict Dataset version 25.1\cite{gleditsch2002armed,pettersson2020organized}. Pearson $r$ is reported separately for OpenAlex, and PubMed.

\subsubsection*{Country-level conflict comparison}

The countries were classified as involved in conflict if cumulative UCDP battle-related deaths (BRD v25.1) during 2010-2024 exceeded 10{,}000, yielding 14 conflict-involved and 30 peaceful countries. Indexed prevalence growth was compared between the two groups (Fig.~\ref{fig:conflict}c), and their post-2019 change in year-on-year slope was compared using a one-sided Mann-Whitney $U$ test (Supplementary Fig.~S3).

\subsubsection*{Conflict-type coupling}
To test whether the conflict association varies by conflict character, active conflicts per year (2010-2024) were counted separately for each UCDP conflict type (interstate, intrastate, and internationalized; unique conflict identifiers). For each type, the yearly count was correlated (Pearson $r$) with the pooled cross-database overall prevalence, and the 15 annual observations were bootstrap-resampled with replacement (2{,}000 draws) to obtain the median correlation and interquartile range (Fig.~\ref{fig:conflict}e).

\subsubsection*{Pre/post-period slope comparison}
The linear regression slopes were estimated separately for pre-period and post-period windows on the pooled cross-database series, with slope ratios used to quantify acceleration. Two split years are used in the paper: 2019 (matching the PELT-detected structural break, used in Fig.~\ref{fig:longitudinal}) and 2020 (matching the COVID-era boundary, used in Fig.~\ref{fig:sensitivity}).

\subsubsection*{COVID sensitivity}
The COVID-related papers were identified by case-insensitive regex matching against title and abstract for the terms: covid, sars-cov-2, coronavirus, pandemic, lockdown, social distancing, and quarantine. The corpus was partitioned into COVID-related and non-COVID papers, and yearly prevalence was re-aggregated separately on each partition.

\subsubsection*{LLM-era language comparison}
The pre-LLM (2010-2022) and post-LLM (2023-2025) linear-regression slopes were estimated separately for native-English-affiliated authors (United States, United Kingdom, Canada, Australia, New Zealand, Ireland) and non-English-affiliated authors. The slope ratios were used to compare the magnitude of acceleration between the language groups.

\subsection*{Survey experiment}

\subsubsection*{Design}
We conducted a within-subject experiment to quantify the perceptual consequences of war framing in scientific abstracts. The abstract pool consisted of 50 short scientific abstracts drawn from five domains (chemistry, cybersecurity, ecology, immunology and microbiology), with 10 abstracts per domain. The five domains were selected to span the corpus disciplinary range while deliberately including biomedical fields (immunology, microbiology) where war metaphors are most documented in prior work\cite{martin1994flexible,penson2004cancer,board2006ending}. The initial abstract drafts were generated using a large language model and then manually edited to ensure factual coherence, parallel scientific content across the war-framed and neutrally-framed versions, and minimal lexical divergence outside the framing manipulation. The stimulus quality was assessed through a manual review of all 50 pairs for content equivalence and parallelism. Each abstract had a war-framed and a neutrally-framed wording (for example, the war-framed version replaced ``address the disease'' with ``combat the disease''), and the framing assignment for each respondent was recorded in a per-respondent randomization map. Each respondent read 40 of the 50 abstracts in a single Qualtrics session, with framing assignment balanced within respondent (20 war-framed, 20 neutrally-framed trials). After each abstract, respondents rated the work on four 7-point Likert scales: perceived credibility, urgency, policy support and willingness to let government fund. The credibility item was adapted from the Münster Epistemic Trustworthiness Inventory\cite{hendriks2015meti}; the remaining three items were single-item measures inspired by outcomes used in prior metaphor-framing research on policy preference\cite{thibodeau2011metaphors}, behavioral intentions under war framing\cite{hauser2015war} and urgency primes from war metaphor\cite{flusberg2017metaphors}.

\subsubsection*{Recruitment and inclusion}
Respondents were recruited as a convenience sample from the US adult Prolific panel\cite{peer2022data} without demographic quotas. We set an initial recruitment target of approximately 1{,}200 respondents based on conventional within-subject framing-experiment sample sizes and an anticipated dropout/exclusion rate of 25--30\%; no formal power analysis was performed \emph{a priori}. The eligible respondents were US-based, aged 18 or older. Of 1{,}159 respondents who completed the Qualtrics session, sessions shorter than 1{,}200 seconds (20 minutes) were excluded as indicating disengaged or straight-lining responses. The final analysis sample comprised $N = 801$ respondents (69\% of completers) and $32{,}040$ trials, with 16{,}020 war-framed and 16{,}020 neutrally-framed trials. Sample demographic characteristics (age, gender, race/ethnicity, education, household income) are reported in Supplementary Table~S1.

\subsubsection*{Effect estimation}

For each outcome, the war-versus-neutral effect was estimated by a linear mixed-effects model with the outcome regressed on the war indicator and crossed random intercepts for respondent and abstract, fitted by restricted maximum likelihood. Treating abstract as a random effect addresses the stimulus-sampling problem and lets inference generalize beyond the specific 50 abstracts used. Within-subject Cohen's $d_z$ was computed from per-respondent paired means as $d_z = \bar{d}/s_d$, where $\bar{d}$ is the mean of within-respondent (war minus neutral) differences and $s_d$ is the standard deviation of those differences\cite{lakens2013calculating}. We report 95\% confidence intervals on the fixed-effect coefficient (Likert units), with effect sizes interpreted against standard benchmarks\cite{cumming2014new,funder2019evaluating}.

\subsubsection*{Demographic moderation}
The subgroup analyses repeated the war-versus-neutral effect estimation within strata of self-reported gender (woman, man), household income ($\leq$\$50K vs $>$\$50K), and educational attainment ($\leq$ some college vs $\geq$ Bachelor's degree). For each moderator, the war $\times$ moderator interaction was tested within the same linear mixed-effects framework as the main effects, regressing the outcome on the war indicator, the moderator indicator, their interaction and crossed random intercepts for respondent and abstract; the interaction coefficient, 95\% confidence interval and $p$-value are reported. The age $\times$ race intersection was chosen \emph{a priori} based on prior evidence that race-based mistrust of scientific and medical institutions interacts with generational cohort effects: race differences in trust in scientific authority have a documented historical basis\cite{corbie2002distrust}, and broader institutional trust shows generational variation that may modulate the race effect across age cohorts\cite{twenge2014generational}. Intersectional age $\times$ race effects were computed on the eight cells formed by $\{<30, \geq30\} \times \{$White, Black, Hispanic/Latino, Asian$\}$.

\subsubsection*{Within-respondent and within-trial associations}
The Pearson correlation between per-respondent (war minus neutral) deltas in credibility and willingness-to-fund was computed across all respondents with both deltas defined. The within-person trial-level coupling between credibility and funding ratings was estimated by two complementary methods. First, a repeated-measures correlation\cite{bakdash2017repeated} was computed by demeaning each respondent's credibility and funding ratings against their respondent-level means and then taking the Pearson correlation across all 32{,}040 trials, with degrees of freedom $df = N - k - 1$ where $N$ is the total trial count and $k$ is the number of respondents; the 95\% confidence interval on $r_{rm}$ was estimated by Fisher $z$-transformation. Second, a linear mixed-effects model with funding as the outcome, credibility as the fixed-effect predictor and crossed random intercepts for respondent and abstract was fitted by restricted maximum likelihood; the fixed-effect slope $\beta$ on credibility, its 95\% confidence interval, and $p$-value are reported alongside the intraclass correlation coefficients (ICCs) for respondent and abstract.

\section*{Citation diversity statement}
Recent work in several fields of science has identified a bias in citation practices such that papers from women and other minority scholars are under-cited relative to the number of such papers in the field \cite{mitchell2013gendered,dion2018gendered,caplar2017quantitative, maliniak2013gender, Dworkin2020.01.03.894378, bertolero2021racial, wang2021gendered, chatterjee2021gender, fulvio2021imbalance}. Here we sought to proactively consider choosing references that reflect the diversity of the field in thought, form of contribution, gender, race, ethnicity, and other factors. First, we obtained the predicted gender of the first and last author of each reference by using databases that store the probability of a first name being carried by a woman \cite{Dworkin2020.01.03.894378,zhou_dale_2020_3672110}. By this measure (and excluding self-citations to the first and last authors of our current paper), our references contain 18.86\% woman(first)/woman(last), 16.23\% man/woman, 12.40\% woman/man, and 52.52\% man/man. This method is limited in that a) names, pronouns, and social media profiles used to construct the databases may not, in every case, be indicative of gender identity and b) it cannot account for intersex, non-binary, or transgender people. Second, we obtained predicted racial/ethnic category of the first and last author of each reference by databases that store the probability of a first and last name being carried by an author of color \cite{ambekar2009name, sood2018predicting}. By this measure (and excluding self-citations), our references contain 8.22\% author of color (first)/author of color(last), 12.58\% white author/author of color, 17.93\% author of color/white author, and 61.27\% white author/white author. This method is limited in that a) names and Florida Voter Data to make the predictions may not be indicative of racial/ethnic identity, and b) it cannot account for Indigenous and mixed-race authors, or those who may face differential biases due to the ambiguous racialization or ethnicization of their names. We look forward to future work that could help us to better understand how to support equitable practices in science.
\pagebreak

\section*{Data availability}
The bibliographic corpora analyzed in this study are publicly available. OpenAlex records were retrieved through the OpenAlex API (\url{https://openalex.org}; CC0 public-domain dedication)\cite{priem2022openalex}, and PubMed records through the NCBI E-utilities API (\url{https://www.ncbi.nlm.nih.gov/pubmed}). Armed-conflict data are publicly available from the Uppsala Conflict Data Program / PRIO Armed Conflict Dataset version 25.1 (\url{https://ucdp.uu.se})\cite{gleditsch2002armed,pettersson2020organized}. The de-identified trial-level survey responses are available from the corresponding authors upon reasonable request. Source data are provided with this paper.

\section*{Code availability}
All codes and statistics required to reproduce the figures are made available at \url{https://www.soveshmohapatra.com/research/warnlp}.

\pagebreak

\bibliography{ref}

\pagebreak

\section*{Acknowledgments}
S.M. and D.S.B. acknowledge support from the Army Research Office MURI program (through grant number W911NF2410228). D.L-S acknowledges support from the National Institute on Drug Abuse (K01 DA047417). The views and conclusions contained in this document are those of the authors and should not be interpreted as representing the official policies of the funders.

\section*{Author Contributions}
Conceptualization: S.M., D.S.B. Methodology: S.M., D.L-S, D.S.B. Data Analyses, Writing -- Original Draft: S.M. Writing -- Review \& Editing: S.M., D.L-S, D.S.B. Supervision: D.S.B.

\section*{Competing Interests}
The authors declare no competing interests.

\end{document}

\maketitle
\pagebreak

\begin{figure}[htbp]
    \centering
    \includegraphics[width=\textwidth]{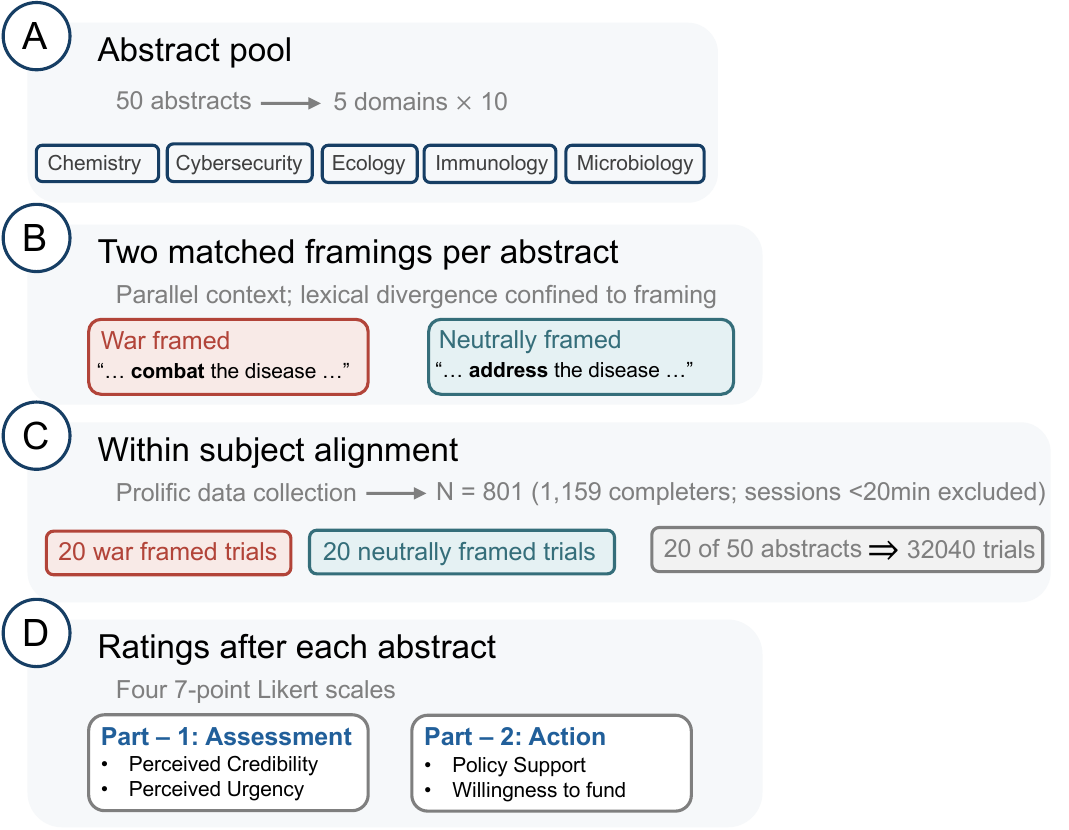}
    \caption{\textbf{Schematic of the within-subject war-framing experiment.}
    (\textbf{A})~Fifty scientific abstracts spanning five domains (10 each).
    (\textbf{B})~Each abstract was written in two content-matched versions differing only in framing (e.g.\ ``combat'' vs.\ ``address'' the disease).
    (\textbf{C})~Each Prolific respondent rated 20 paired of the 50 paired abstracts, balanced within respondent (20 war-framed, 20 neutral); after excluding sessions under 20 minutes, $N = 801$ of 1{,}159 completers yielded $32{,}040$ trials.
    (\textbf{D})~After each abstract, respondents gave four 7-point Likert ratings: perceived credibility and urgency (assessment), policy support and willingness to fund (action).}
    \label{fig:design}
\end{figure}

\begin{table}[htbp]
\centering
\caption{\textbf{Demographic characteristics of the survey analysis sample ($N = 801$).} Percentages are of the total analysis sample. Age was self-reported and 5 respondents (0.6\%) entered implausible values that were excluded from the age statistic; bucketed age categories are computed from the cleaned values, with implausibly large entries floored into the $\geq 60$ category.}
\label{tab:demographics}
\begin{tabular}{lrr}
\toprule
Characteristic & $n$ & \% \\
\midrule
\textbf{Age (years)} & & \\
\quad Mean (SD) & \multicolumn{2}{r}{38.2 (9.5)} \\
\quad Median & \multicolumn{2}{r}{37} \\
\quad Range & \multicolumn{2}{r}{18--60} \\
\quad Valid age reported & 796 & 99.4 \\
\addlinespace
\textit{Age band} & & \\
\quad 18--29 & 156 & 19.5 \\
\quad 30--44 & 438 & 54.7 \\
\quad 45--59 & 189 & 23.6 \\
\quad $\geq$ 60 & 18 & 2.2 \\
\addlinespace
\textbf{Gender} & & \\
\quad Woman & 456 & 56.9 \\
\quad Man & 329 & 41.1 \\
\quad Nonbinary / other & 16 & 2.0 \\
\addlinespace
\textbf{Race / ethnicity} & & \\
\quad White & 485 & 60.5 \\
\quad Black & 191 & 23.8 \\
\quad Hispanic / Latino & 69 & 8.6 \\
\quad Asian & 40 & 5.0 \\
\quad Native / Indigenous & 9 & 1.1 \\
\quad Multiracial & 1 & 0.1 \\
\quad Other & 6 & 0.7 \\
\addlinespace
\textbf{Educational attainment} & & \\
\quad $\leq$ High school & 193 & 24.1 \\
\quad Some college & 113 & 14.1 \\
\quad Bachelor's degree & 317 & 39.6 \\
\quad Graduate degree & 174 & 21.7 \\
\quad Not reported & 4 & 0.5 \\
\addlinespace
\textbf{Household income} & & \\
\quad $<$ \$25k & 101 & 12.6 \\
\quad \$25k--50k & 160 & 20.0 \\
\quad \$50k--75k & 159 & 19.9 \\
\quad \$75k--100k & 130 & 16.2 \\
\quad $>$ \$100k & 225 & 28.1 \\
\quad Not reported & 26 & 3.2 \\
\bottomrule
\end{tabular}
\end{table}

\clearpage
% -- Lexicon & corpus --
\begin{table}[htbp]
\centering
\caption{\textbf{The three-tier militarized-language lexicon.} All 141 base terms used by the counting pipeline, grouped by tier. Matching is case-insensitive with word boundaries and regular-plural expansion (e.g.\ \texttt{weapon} matches \texttt{weapons}); irregular plurals are matched explicitly. Tier~3 matches were additionally disambiguated by a large-language-model classifier to remove non-militarized scientific usages (Methods).}
\label{tab:lexicon}
\small
\begin{tabular}{p{0.15\textwidth}p{0.76\textwidth}}
\toprule
Tier & Terms \\
\midrule
\textbf{Tier 1}\newline Literal military\newline (51 terms) & war, warfare, military, militarization, militarized, army, navy, airforce, battlefield, battleground, combatant, troop, infantry, artillery, armament, munition, bloodshed, warplane, warship, ceasefire, armistice, peacekeeping, terrorism, terrorist, militant, paramilitary, guerrilla, mercenary, sniper, platoon, battalion, regiment, brigade, commander, lieutenant, general, admiral, pentagon, insurgency, insurgent, jihad, kamikaze, homeland security, counterinsurgency, counterterrorism, demilitarized, noncombatant, warlord, blitzkrieg, wmd, weapons of mass destruction \\
\addlinespace
\textbf{Tier 2}\newline Combat metaphor\newline (46 terms) & battle, combat, weapon, weaponize, weaponized, invade, invasion, invader, enemy, adversary, ally, alliance, surrender, retreat, bombard, bombardment, ambush, raid, casualty, hostage, civilian, refugee, veteran, militia, siege, arms race, escalate, escalation, retaliation, retaliate, preemptive, hostility, hostile, frontline, blockade, espionage, sabotage, tactic, tactical, maneuver, mobilize, mobilization, demobilize, collateral damage, cyberwarfare, cyberattack \\
\addlinespace
\textbf{Tier 3}\newline Ambiguous dual-use\newline (44 terms) & attack, defense, defend, target, strategy, deploy, deployment, campaign, strike, shield, intercept, conflict, resist, resistance, guard, trigger, threat, conquer, conquest, defeat, victory, survivor, survive, protect, protection, secure, security, vulnerability, vulnerable, aggressive, aggression, force, operation, mission, division, command, base, camp, recruit, recruitment, drill, uniform, rank, intelligence \\
\bottomrule
\end{tabular}
\end{table}

\clearpage
% -- Fig 2 support --
\begin{figure}[htbp]
\centering
\includegraphics[width=0.55\textwidth]{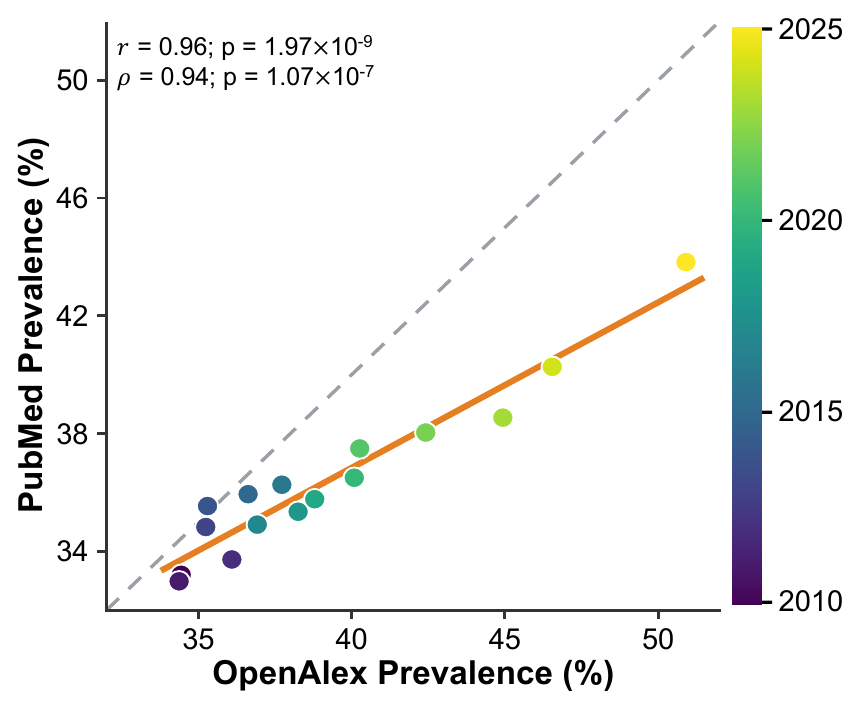}
\caption{\textbf{Cross-database concordance.} Annual overall militarized-language prevalence in OpenAlex (x-axis) versus PubMed (y-axis), 2010-2025 (points colored by year, dark = 2010 to bright = 2025). The two independent corpora track each other closely (Pearson $r = 0.96$, $p < 10^{-8}$; Spearman $\rho = 0.94$) and sit near the identity line (dashed), indicating the trend is not an artifact of either database. Solid line: OLS fit.}
\label{fig:concordance}
\end{figure}

\begin{table}[htbp]
\centering
\caption{\textbf{Segmented (change-point) regression of tier prevalence on year, break at 2019.} Slopes are percentage points per year with 95\% CIs; $\Delta$slope $p$ tests whether the post-2019 slope differs from pre-2019. Tier~3 shows a $\sim$4.6-fold acceleration; Tier~1 reverses to a decline; Tier~2 is unchanged.}
\label{tab:segmented}
\begin{tabular}{lcccc}
\toprule
Tier & Pre-2019 slope [95\% CI] & Post-2019 slope [95\% CI] & $\Delta$slope $p$ & $R^2$ \\
\midrule
Tier 1 & +0.02 [-0.06, +0.08] & -0.14 [-0.24, -0.04] & 0.018 & 0.44 \\
Tier 2 & +0.05 [+0.04, +0.07] & +0.07 [+0.04, +0.09] & 0.28 & 0.95 \\
Tier 3 & +0.37 [+0.12, +0.62] & +1.70 [+1.33, +2.06] & $3\times10^{-5}$ & 0.96 \\
\bottomrule
\end{tabular}
\end{table}

\clearpage
% -- Fig 5 support --
\begin{figure}[htbp]
\centering
\includegraphics[width=0.5\textwidth]{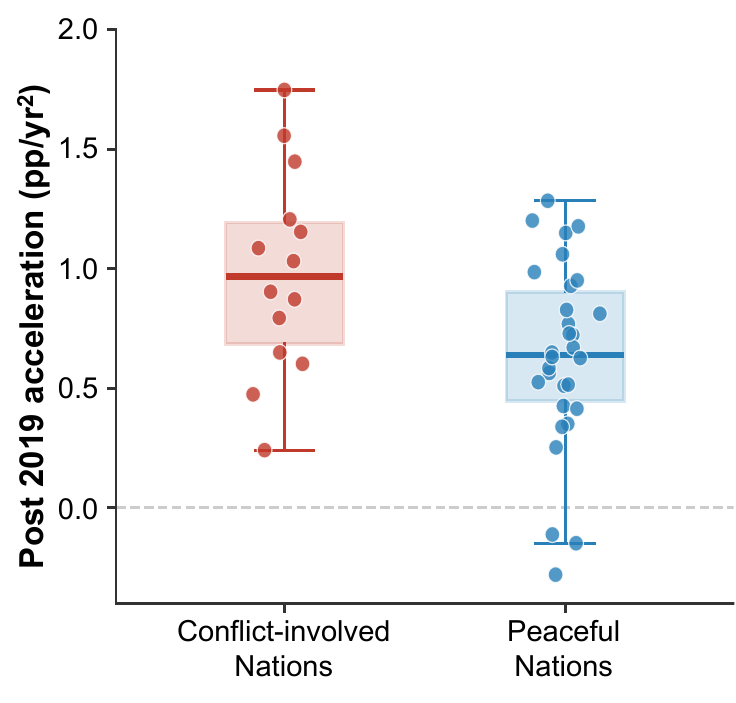}
\caption{\textbf{Conflict-involved nations accelerate faster than peaceful nations.} Post-2019 change in the year-on-year slope of overall militarized-language prevalence (percentage points per year$^2$; pooled across both databases) for each country, split by conflict involvement (conflict-involved, crimson, $n = 14$; peaceful, blue, $n = 30$). Boxes show the interquartile range and median; points are individual countries. Conflict-involved nations accelerate more steeply after 2019 (median $+0.97$ vs $+0.64$~pp\,yr$^{-2}$; one-sided Mann--Whitney $U$, $p = 0.011$), a differential response that complements the aggregate association in Fig.~5A--C and the country-level case studies in Fig.~5D.}
\label{fig:accel}
\end{figure}

\clearpage
% -- Fig 6 support --
\begin{figure}[htbp]
\centering
\includegraphics[width=0.6\textwidth]{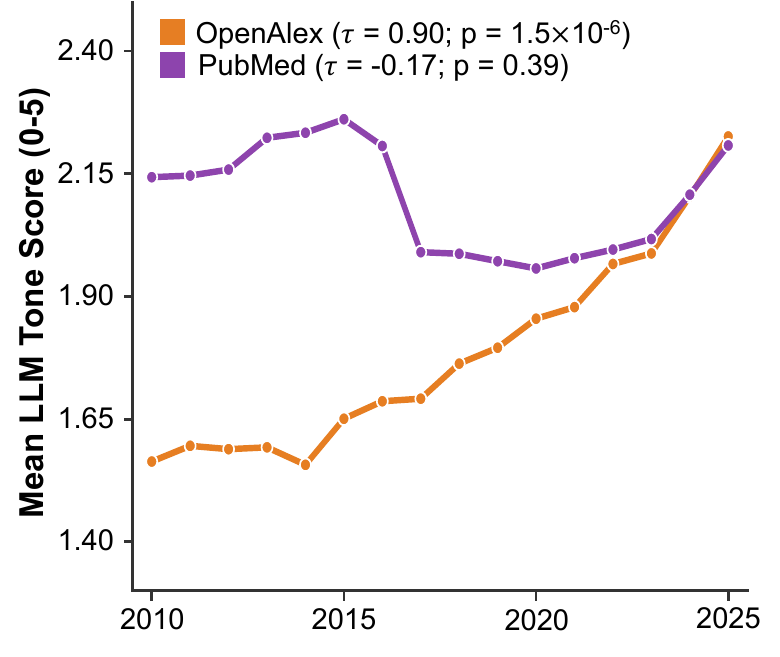}
\caption{\textbf{Independent large-language-model tone validation, by database.} Mean LLM-assigned militarization tone (0--5; Methods) per year (x-axis, 2010-2025; y-axis, mean tone). OpenAlex in orange, PubMed in purple. In OpenAlex tone rises in lockstep with the lexicon signal (Mann--Kendall $\tau = 0.90$, $p < 10^{-5}$; tone-vs-lexicon $r = 0.98$), giving convergent validation. In PubMed tone shows no significant trend (Mann--Kendall $\tau = -0.17$, $p = 0.39$; tone-vs-lexicon $r = -0.03$), consistent with PubMed's weaker lexicon trend; this database inconsistency is why tone is reported here rather than in the main text.}
\label{fig:tone}
\end{figure}

\clearpage
% -- Fig 7 support --
\begin{figure}[htbp]
\centering
\includegraphics[width=0.6\textwidth]{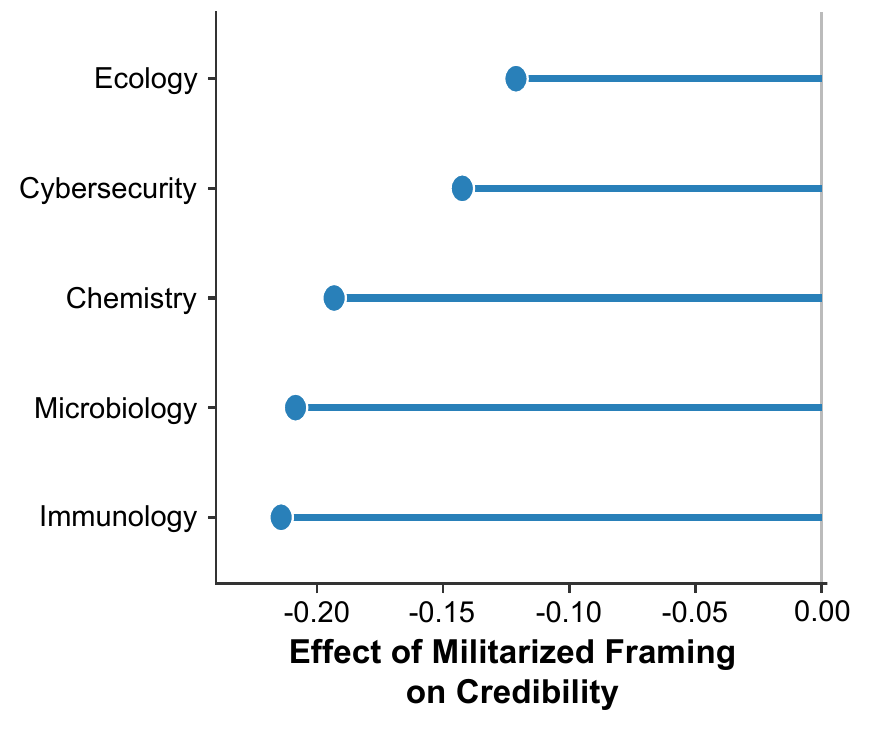}
\caption{\textbf{The credibility penalty generalizes across scientific domains.} Within-subject effect of militarized framing on perceived credibility ($d_z$; x-axis) for each of the five survey domains (rows, top to bottom: ecology, cybersecurity, chemistry, microbiology, immunology). The penalty is negative and significant in every domain (immunology and microbiology $d_z = -0.21$; chemistry $-0.19$; cybersecurity $-0.14$; ecology $-0.12$; all $p < 10^{-3}$), showing the main-text effect (Fig.~7A) is not driven by any single field.}
\label{fig:domain}
\end{figure}

\begin{table}[htbp]
\centering
\caption{\textbf{Construct validity: correlations among the four outcome effects.} Pearson correlations between respondent-level war-minus-neutral deltas for the four outcomes ($N = 801$). The assessment measures (credibility, urgency) and action measures (policy support, funding) form coherent but distinct constructs; policy support and funding are most strongly related ($r = 0.81$), while urgency is the most distinct ($r \leq 0.50$ with all others).}
\label{tab:construct}
\begin{tabular}{lcccc}
\toprule
 & Credibility & Urgency & Policy support & Gov. funding \\
\midrule
Credibility & 1.00 &  &  &  \\
Urgency & 0.38 & 1.00 &  &  \\
Policy support & 0.66 & 0.48 & 1.00 &  \\
Gov. funding & 0.63 & 0.50 & 0.81 & 1.00 \\
\bottomrule
\end{tabular}
\end{table}

\begin{table}[htbp]
\centering
\caption{\textbf{Response-quality and design-balance checks.} Summary of the survey validity battery. The duration filter is the only exclusion; the resulting sample is fully balanced by design.}
\label{tab:quality}
\begin{tabular}{lr}
\toprule
Check & Value \\
\midrule
Completers & 1{,}159 \\
Analysis sample (session $\geq$ 20 min) & 801 (69\%) \\
Trials per respondent & 40 (all) \\
War / neutral balance per respondent & 0.50 (SD $<$ 0.001) \\
War-framed trials analyzed & 16{,}020 \\
Neutral trials analyzed & 16{,}020 \\
\bottomrule
\end{tabular}
\end{table}